\documentclass[a4paper,conference]{IEEEtran}

\usepackage[utf8]{inputenc} 
\usepackage[T1]{fontenc}    
\usepackage{url}            
\usepackage{booktabs}       
\usepackage{amsfonts}       
\usepackage{nicefrac}       
\usepackage{microtype}      
\usepackage{times}
\usepackage{amsmath,bm}
\usepackage{amssymb}
\usepackage{centernot}
\usepackage{algorithm}
\usepackage{algorithmic}
\usepackage{float}
\usepackage{wrapfig}
\usepackage{parskip}
\usepackage{multicol}
\usepackage{cuted}
\usepackage{ulem}
\usepackage{cancel}

\usepackage{ifthen}
\usepackage{fancyhdr}
\renewcommand{\headrulewidth}{0pt}
\renewcommand{\footrulewidth}{0pt}

\usepackage{times,amsmath,epsfig,graphicx,amssymb,multirow,subfig}

\def\ie{\textit{i}.\textit{e}., }

\usepackage{mathtools,amssymb}     
\usepackage{tikz}
\usepackage{xparse}
\NewDocumentCommand{\xleftrightarrows}{ O{}O{} }{%
\mathrel{%
\vcenter{\hbox{%
\begin{tikzpicture}
  \node[minimum width=1cm,minimum height=1ex,anchor=south,align=center] (a){\text{\vphantom{hg}#1}\\[0.5ex] \vphantom{hg}#2};
  \draw[->] ([yshift=0.35ex]a.west) -- ([yshift=0.35ex]a.east);
  \draw[<-] ([yshift=-0.35ex]a.west) -- ([yshift=-0.35ex]a.east);
\end{tikzpicture}
}}%
}%
}

\NewDocumentCommand\DownArrow{O{2.0ex} O{black}}{%
   \mathrel{\tikz[baseline] \draw [<-, line width=0.5pt, #2] (0,-0.75em) -- ++(0,#1);}
}
\NewDocumentCommand\UpArrow{O{2.0ex} O{black}}{%
   \mathrel{\tikz[baseline] \draw [->, line width=0.5pt, #2] (0,-0.75em) -- ++(0,#1);}
}

\newtheorem{definition}{Definition}

\def\Reals{\mathbb{R}}

\def\th{^{\mbox{\small th}}}

\def\vec#1{{\mathbf{#1}}}
\def\mat#1{{\mathbf{#1}}}

\def\ten#1{{\mathbf{\mathcal#1}}}

\def\tp{^{\mathrm{T}}}

\def\oast{\circledast}

\def\assign{\mathrel{\mathop:}=}

\def\mode#1{_{\mbox{\tiny\textrm{#1}}}}
\def\tmode#1{_{\mbox{\tiny$\ten{#1}$}}}

\def\measure{\mode 0}
\def\pixels{\mode x}

\def\people{\mode P}
\def\views{\mode V}
\def\illums{\mode L}

\def\matize#1#2{{\mat#1}\mode{$[#2]$}}

\def\T{{}^{\hbox{\tiny\textrm{T}}}}

\def\inv{^{-1\lower2pt\hbox{\hskip-2pt\hbox{\tiny${}$}}}}
\def\pinv#1{^{+\lower2pt\hbox{\hskip-2pt\hbox{\tiny${}#1$}}}}

\usepackage[pagebackref=false,breaklinks=true,colorlinks,bookmarks=false]{hyperref}

\ifCLASSINFOpdf
\else
\fi
\usepackage{algorithmic}
\usepackage{fixltx2e}

\usepackage{stfloats}
\usepackage{url}

\usepackage[a4paper,top=54pt,bottom=80pt,left=37pt,right=37pt]{geometry}

\renewcommand{\baselinestretch}{.95}
\begin{document}
\title{\vskip-.05in
CausalX: Causal\hspace{+.0in} eXplanations\hspace{+.0in} \\and 
Block Multilinear Factor Analysis
\vskip-.05in
}

%
\title{\vskip-.05in
CausalX: 
Causal\hspace{+.0in} eXplanations\hspace{+.0in} \\ and 
Block Multilinear Factor Analysis
\vskip-.05in
}

\author{%
\hskip-0in
\IEEEauthorblockN{M. Alex O. Vasilescu$^{1,2}$}
\IEEEauthorblockA{ 
\hskip-0in
maov@cs.ucla.edu}
\and
\IEEEauthorblockN{}
\IEEEauthorblockA{\\
\hskip-.1in 
$^{1}$Tensor Vision Technologies, 
Los Angeles, California 
}
\and
\hskip-3.5in
\IEEEauthorblockN{Eric Kim$^{2,1}$}
\IEEEauthorblockA{
\hskip-3.5in
ekim@cs.ucla.edu}
\and
\IEEEauthorblockN{}
\IEEEauthorblockA{\\
\\
\hskip-4in
$^{2}$Department of Computer Science,
University of California, Los Angeles
\vspace{-.05in}
}
\and
\hskip-.5in
\IEEEauthorblockN{
Xiao S. Zeng$^{2}$} 

\IEEEauthorblockA{
\hskip-.5in
stevennz@ucla.edu
}
\vspace{-.2in}
}


%


\maketitle
\thispagestyle{fancy}
\begin{abstract}
By adhering to the dictum, ``No causation without manipulation (treatment, intervention)'', 
cause and effect data analysis represents changes in observed data in terms of changes in the causal  factors. 
When causal factors are not amenable for active manipulation in the real world due to current technological limitations or ethical considerations, a counterfactual approach
performs 
an intervention on the model of data formation. In the case of object representation or activity (temporal object) representation, varying object parts is generally unfeasible whether they be spatial and/or temporal. 
Multilinear algebra, the algebra of higher order tensors, is a suitable and transparent framework for disentangling the causal factors of data formation. Learning 
a part-based intrinsic causal factor representations in a multilinear framework requires applying a set of interventions on 
a 
part-based multilinear model. We propose a unified 
multilinear model of wholes and parts. We derive a 
hierarchical block multilinear 
factorization, the $M$-mode Block SVD, that computes a disentangled representation of the causal factors by optimizing simultaneously across the entire object 
hierarchy. Given computational efficiency considerations, 
we introduce an incremental bottom-up computational alternative, the Incremental $M$-mode Block SVD, 
that employs the lower level abstractions, the part representations, to represent the higher level of abstractions, the parent wholes. This incremental computational approach may also be employed to update the causal model parameters when data becomes available incrementally. The resulting object representation is an interpretable combinatorial choice of intrinsic causal factor representations related to an object’s recursive 
hierarchy 
of wholes and parts that renders object recognition robust to occlusion and reduces training data requirements.
\vspace{+.075in}
\end{abstract}

\begin{figure*}[!b]
\vskip-.185in
\centering
  \includegraphics[width=\textwidth]{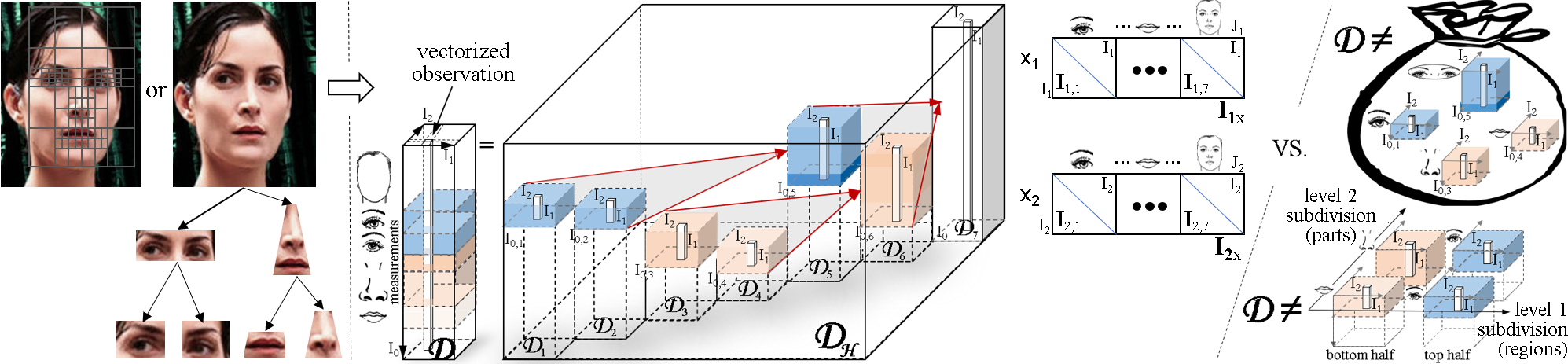}
  \vskip-.015in
  \caption{Data tensor, $\ten D$, expressed in terms of a hierarchical data tensor, $\ten D\tmode H$, a mathematical instantiation of a tree data structure where  $\ten D = \ten D\tmode H \times\mode 1 \mat I\mode{1x} \dots \times\mode c \mat I\mode{cx} \dots \mat I\mode{Cx}$, versus a bag of independent parts/sub-parts, 
  or a data tensor with a reparameterized measurement mode in terms of regions and sub-regions. 
  An object hierarchy may be based on adaptive quad/triangle based subdivision of various depths~\cite{vasilescu92}, or a set of perceptual parts of arbitrary shape, size and location. 
 Images of non-articulated objects 
 are best expressed with 
  hierarchical data tensors that have a partially compositional form, where all the parts share the same extrinsic causal factor representations, Fig.~\ref{fig:Block-Tucker-Base-Case}b. Images of objects with articulated parts are best expressed in terms of 
  hierarchical data tensors that are fully compositional in the causal factors, Fig.~\ref{fig:Block-Tucker-Base-Case}c. Images of non-articulated objects may also be
represented by a fully compositional hierarchical data tensor, as depicted by the TensorTrinity example above. 
  }
 \label{fig:toyexample}
\end{figure*}


\section{Introduction: Problem Definition}
\label{sec:Introduction}
\vspace{-.065in}
Developing 
causal explanations 
for correct results or for failures from mathematical equations and data is important in developing a trustworthy artificial intelligence, and retaining public trust. 
Causal explanations 
are germane 
to the ``right to an explanation'' statute~\cite{Goodman2017,Gilpin18} 
\ie to data driven decisions, such as those that rely on images.
Computer graphics and computer vision problems, also known as forward and inverse imaging problems, have been cast as causal inference questions~\cite{Vasilescu09,Vasilescu19} 
consistent with Donald Rubin's 
quantitative 
definition 
of causality, where ``A causes B'' means ``the effect of A is B'', a measurable and experimentally repeatable quantity
~\cite{Glymour86,Holland86B}. 
Computer graphics may be viewed as addressing analogous questions to forward causal inferencing that addresses the ``what if'' question, and estimates the change in effects g\underline{iven} a delta change in a causal factor. Computer vision 
may be viewed as addressing analogous questions to inverse causal inferencing 
that addresses the 
``why'' question~\cite{gelman13}. 
We define inverse causal inference 
as the estimation of 
causes 
g\underline{iven} an estimated forward causal model and a set of observations that constrain the solution set.

\begin{figure*}[!t]
\begin{tabular}{cc}
  \hskip-.25in
    \begin{tabular}{c}
    \vspace{-.225in}
    \\
        \hskip+.0in
        \psfig{figure=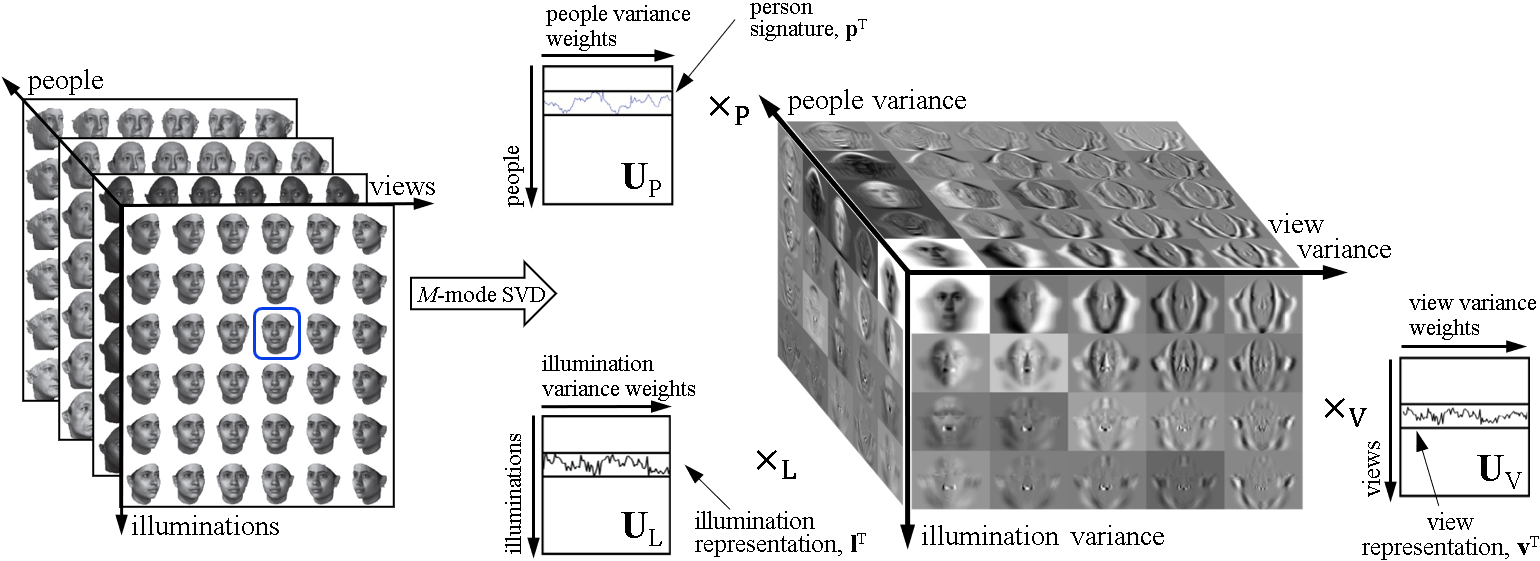,width=.69\linewidth}\\
\vspace{-.175in}
       \mbox{\scriptsize{(a)}}\\
    \end{tabular}
    \hskip-.265in
    \vspace{-.25in}
    \begin{tabular}{c}
       \vspace{-.05in}
       \hspace{-.025in}
       \psfig{figure=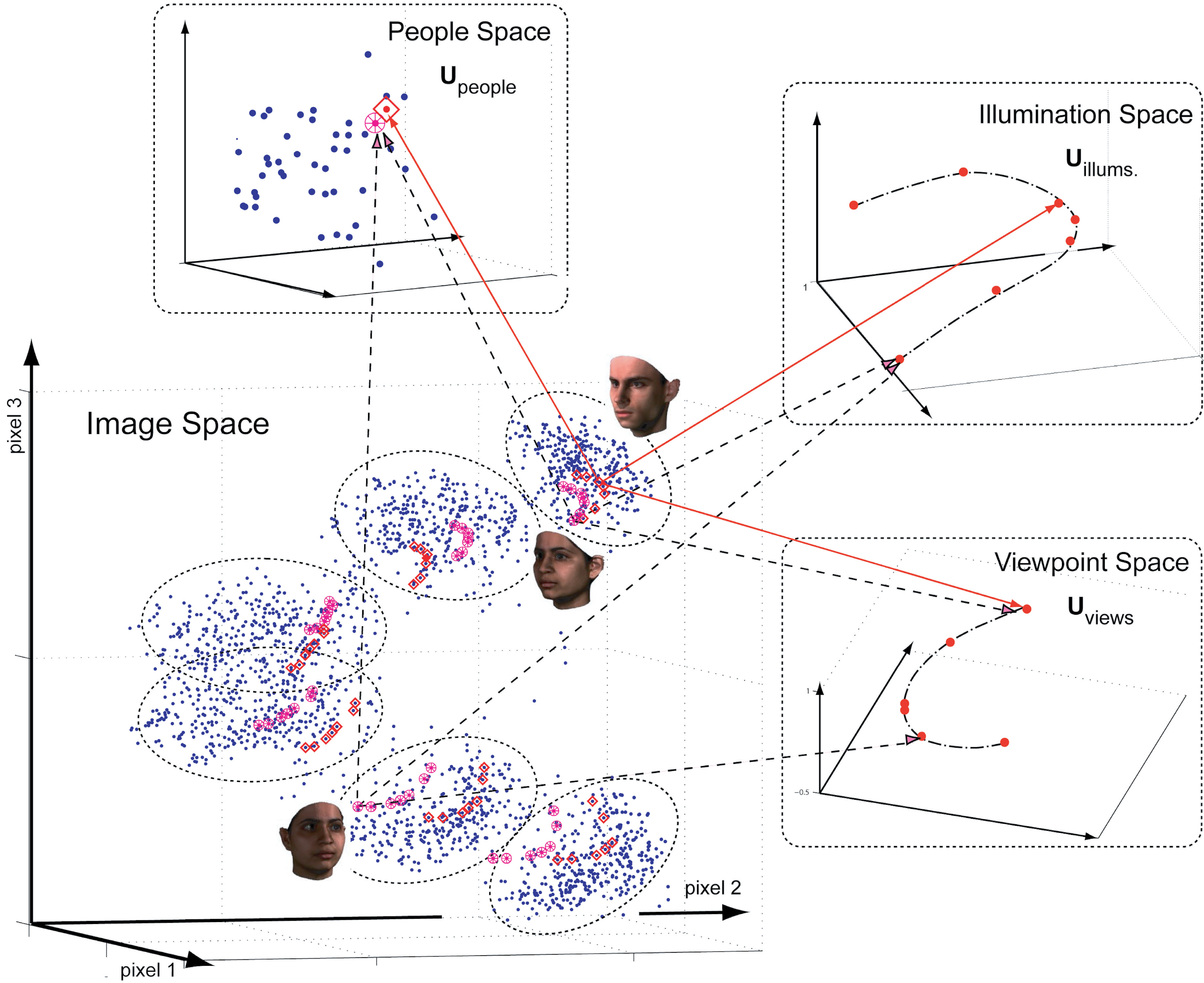,width=.32\linewidth}
       \\
       \vspace{-.15in}
       \\
       \hskip+.1in
       \mbox{\scriptsize{(b)}}\\
    \end{tabular}
    \\
\end{tabular}

\vspace{+.2in}
\caption[MPCA image representation and coefficient vectors]
{(a) Forward Causal Inference: 
A vectorized image, $\vec d$,
is represented by a set of coefficient vectors, one for the illumination, the viewing conditions, and the person ($\vec l, \vec v, \vec p$), and expressed mathematically by $\vec d=\ten T\times\illums \vec l\tp \times\views \vec v\tp\times\people \vec p\tp$.
 The TensorFaces basis, $\ten T$, governs the interaction between the causal factors of data formation~\cite{Vasilescu05}. 
(For display only, the mean was added back.)
(b) Inverse Causal Inference: While TensorFaces (Multilinear-PCA or Multilinear-ICA~\cite{Vasilescu05}) learns the interaction between the causal factors from training data, it does not prescribe an approach for estimating the causal factors from one or  more unlabeled test images that need to be enrolled or recognized.  
For an unlabeled
vectorized test image $\vec d\mode{new}$, 
the causal factor labels are estimated through a multilinear projection algorithm~\cite{Vasilescu07a, Vasilescu2011} that 
is succinctly expressed as the $\mbox{M-mode SVD/CP}\left(\ten{T}\pinv{x} 
\times\pixels\tp {\vec d}\mode{new}\right)\approx \vec r\illums \circ \vec r\views \circ \vec r\people$ where $\vec r\illums$, $\vec r\views$, and $\vec r\people$, are the estimated latent representations from which the view, illumination and person label may be inferred.
}
\label{fig:tensorfaces}
\vspace{-.175in}
\end{figure*}

Natural images are the compositional  
consequence of multiple factors related to scene structure, 
 illumination conditions, and imaging conditions. 
Multilinear algebra, the algebra of higher-order tensors, offers a potent mathematical framework for analyzing the multifactor structure of image ensembles and for addressing the difficult problem of disentangling the constituent factors, Fig.~\ref{fig:tensorfaces}. (Vasilescu and Terzopoulos:
TensorFaces $2002$~\cite{Vasilescu02,Vasilescu03}, 
MPCA and MICA $2005$~\cite{Vasilescu05}, kernel variants~\cite{Vasilescu09}, Multilinear 
Projection $2007/2011$\cite{Vasilescu07a,Vasilescu2011})

\newpage
Scene structure is composed from a set of objects that appear to be formed from a recursive 
hierarchy of perceptual wholes and parts whose properties, such as shape, reflectance, and color, 
constitute a 
hierarchy of intrinsic causal factors of object appearance.
Object appearance is the 
compositional consequence of both an object's intrinsic causal factors, and  extrinsic causal factors with the latter related to 
illumination (i.e. the location and types of light sources),
  and imaging (i.e. viewing direction, camera lens, rendering style etc.). 
 Intrinsic and extrinsic causal factors confound each other's contributions hindering recognition~\cite{Vasilescu19}.

``Intrinsic properties are by virtue of the thing itself and nothing else'' (David Lewis, 1983~\cite{Lewis1983a}); whereas extrinsic properties are not entirely about that thing, but as a result of the way the thing interacts with the world.  
Unlike global intrinsic properties, local intrinsic properties are intrinsic to a part of the thing, and it may be said that a local intrinsic property is in an ``intrinsic fashion'', or ``intrinsically'' about the thing, rather than ``is intrinsic'' to the thing~\cite{Humberstone1996}.

Cause and effect analysis models the mechanisms of data formation, unlike conventional statistical analysis and conventional machine learning that model the distribution of the data~\cite{pearl00}. 
Causal modeling from observational studies 
are suspect of bias and confounding 
with some exceptions~\cite{Cochran72,spirtes00}, unlike experimental studies~\cite{Rubin74, Rubin75} in which a set of active interventions 
are applied, and their effect on response variables are measured and modeled. The differences between experimental studies, denoted symbolically with Judea Pearl's $do$-operator~\cite{pearl00}, and observational studies 
are best exemplified by the following expectation and probability expressions
\begin{eqnarray}
\hspace{-.05in}\begin{array}{ccc}
\hskip+.2in 
E(\vec d | c) 
&\ne& \hskip+.2in 
E(\vec d | do(c))\\
\hskip+.2in \underbrace{P(\vec d | c)}
&\ne& \hskip+.25in 
\underbrace{P(\vec d | do(c))},\\
\hskip+.2in \mbox{\scriptsize \bf From Observational Studies:}  
& &  
\hskip+.2in 
\mbox{\scriptsize \bf From Experimental Studies:}\\
\hskip+.2in 
\mbox{\scriptsize Association, Correlation, Prediction}  
& &  \hskip+.2in 
\mbox{\scriptsize Causation}\\
\end{array}\nonumber
\end{eqnarray}
\vskip-.15in
\noindent
where $\vec d$ is a multivariate observation, and $c$ is a hypothesized or actual causal factor. 
Pearl and Bareinboim~\cite{Pearl14,Bareinboim2016} have delineated the challenges of generalizing results from 
experimental studies 
to observational studies 
by parameterizing the error based on the possible error inducing sources.

The multilinear (tensor) structural equation approach is a suitable and transparent framework for disentangling 
 the factors of data formation  
that has been 
employed in psychometrics~\cite{Tucker66,Harshman70,Carroll70,Bentler79} 
 econometrics~\cite{Magnus88},
 chemometrics~\cite{Bro97,acar2014}, signal processing~\cite{Delathauwer97,Delathauwer08b,Lim14,Markopoulos18}
computer vision
~\cite{Vasilescu03,Elgammal04,Wang03b,wang17}, computer graphics~\cite{Vasilescu02a, Vasilescu04,Vlasic05,Hsu05,Liu11,Miandji19},
 and machine learning
\cite{Vasilescu09,Vasilescu05,chu09,tang13}. 

Adhering to the dictum, ``No causation without manipulation (treatment, intervention)''~\cite{Rubin75,Imbens2015} 
each  causal factor is varied one at a time while holding the rest fixed, and their effects on the response variables are measured and modeled by a data tensor model. The best evidence comes from randomized comparative studies. 
However, when causal factors are not amenable for manipulation due to current technological limitations or ethical considerations, a counterfactual approach is required. Rather than performing a manipulation in the real world, a counterfactual approach performs an intervention on the model.

In the case of object representation or activity (temporal object) representation, varying object parts is generally unfeasible whether they be spatial or temporal. Learning a 
hierarchy of intrinsic causal factor representations requires applying a set of interventions on the structural model, hence it requires a part-based 
multilinear model, Fig~\ref{fig:toyexample}.

This paper 
proposes a unified 
multilinear model of wholes and parts that defines a data tensor 
in terms of 
a {\it hierarchical data tensor},
$\ten D\tmode H$, a mathematical instantiation of a tree data structure.
Our hierarchical data tensor is a mathematical conceptual device 
that 
allows for a different tree parameterization for each causal factor, and
enables 
us to derive a 
multilinear hierarchical block factorization, an $M$-mode Block SVD, that optimizes simultaneously across the entire object hierarchy. 
Given computational considerations, we develop an incremental computational alternative that employs the lower level abstractions, the part representations, to represent the higher level of abstractions, the parent wholes. 

Our hierarchical block multilinear factorization, $M$-mode Block SVD, disentangles the causal
structure by computing statistically invariant intrinsic and extrinsic representations. The factorization learns a hierarchy of low-level, mid-level and high-level features. Our hybrid approach mitigates the shortcomings of local features that are sensitive to local deformations and noise, and the shortcomings of global features that are sensitive to occlusions. The resulting object representation is a combinatorial choice of part representations, that renders object recognition robust to occlusion and reduces large training data requirements.  
This approach was employed for face verification by computing a set of causal explanations (causalX)~\cite{Vasilescu19}. 

\vspace{+.025in}
\section{Relevant Tensor Algebra}
\vspace{-.05in}
{We will use standard textbook notation, denoting scalars by lower case italic letters $(\it{a, b, ...})$, vectors by bold lower case letters $(\vec{a, b, ...})$, matrices by bold uppercase letters $(\mat A, \mat B,...)$, and higher-order tensors by bold uppercase calligraphic letters $(\ten A, \ten B,...)$. Index upper bounds are denoted by italic uppercase (\ie  $1\le i \le I$).  The zero matrix is denoted by $\mat 0$, and the identity matrix is denoted by $\mat I$. References~\cite{Kolda09,Sidiropoulo2017} provide a quick tutorial, but references~\cite{Vasilescu09,Vasilescu05,Vasilescu2011} are an indepth treatment of tensor based factor analysis.
}


Briefly, the natural generalization of matrices (i.e., linear operators defined
over a vector space), tensors define multilinear operators over a {\it
set} of vector spaces. 
A {\it ``data tensor''} denotes an $M$-way data array. 

\begin{definition}[Tensor]
Tensors are multilinear mappings over a set of vector spaces, $\Reals^{I_c}$, $1\le c\le C$, to a range vector space $\Reals^{I_0}$:
\vspace{-.05in}
\begin{equation} 
\ten A: \left\{\Reals^{I_1} \times \Reals^{I_2} \times \dots \times
\Reals^{I_C}\right\} \mapsto \Reals^{I_0}.
\end{equation}
The {\it
order} of tensor $\ten A \in \Reals^{I_0 \times I_1 \times \dots
\times I_C}$ is $M=C+1$. An element of $\ten A$ is denoted as $\ten A_{i_0 i_1
\dots i_c \dots i_C}$ or $a_{i_0 i_1\dots i_c \dots i_C}$, where $1\le i_0\le I_0$, and $1\le i_c \le I_c$.
\end{definition}
The mode-$m$ vectors of an $M\th$-order tensor $\ten A\in\Reals^{I_1
\times I_2 \times \dots \times I_M}$ are the $I_m$-dimensional vectors
obtained from $\ten A$ by varying index $i_m$ while keeping the other
indices fixed. In tensor terminology, column vectors are the mode-1
vectors and row vectors as mode-2 vectors. 
The mode-$m$ vectors of a tensor are also known as {\it fibers}. The
mode-$m$ vectors are the column vectors of matrix $\matize A m$ that
results from {\it matrixizing} (a.k.a. {\it flattening}) the tensor
$\ten A$. 
\vspace{-0.025in}
\begin{definition}[Mode-$m$ Matrixizing]
The mode-$m$ matrixizing of tensor $\ten A \in \Reals^{I_1\times
I_2\times \dots I_M}$ is defined as the matrix $\matize A m \in
\Reals^{I_m \times (I_1 \dots I_{m-1} I_{m+1} \dots I_M)}$.
As the parenthetical ordering indicates, the mode-$m$ column vectors
are arranged by sweeping all the other mode indices through their 
ranges, with smaller mode indexes varying more rapidly than larger
ones; thus,
\vspace{-.025in}
\begin{eqnarray}
&&\left[\matize A m\right]_{jk} 
\hskip-.05in= a_{i_1\dots i_m\dots i_M},
\quad\mbox{where}
\hskip-.2in
\\
&&\hspace{+.5in}\quad j=i_m \quad \mbox{and} \quad k=1+\sum_{n=0\atop
n\neq m}^M(i_n - 1)\prod_{l=0\atop l\neq m}^{n-1} I_l.\nonumber
\label{eq:matrixizing}
\vspace{-.2in}
\end{eqnarray}
\label{def:matrixizing}
\vspace{-.25in}
\end{definition}
\noindent
A generalization of the product of two matrices is the product of a
tensor and a matrix \cite{Delathauwer97}.
\begin{definition}[Mode-$m$ Product, $\times\mode m$]
\label{def:mode-m-prod}
The mode-$m$ product of a tensor 
$\ten A \in \Reals^{I_1 \times I_2 \times \dots \times I_m \times \dots \times I_M}$ 
and a matrix 
$\mat B \in \Reals^{J_m \times I_m}$, 
denoted by $\ten A \times_m \mat B$, is
a tensor of dimensionality 
$\Reals^{I_1 \times \dots \times I_{m-1}
\times J_m \times I_{m+1} \times \dots \times I_M}$ whose entries are
computed by
\begin{eqnarray}
\vspace{-.075in}
[\ten A \times_m \mat B]_{i_1 \dots i_{m-1} j_m i_{m+1} \dots i_M}\hspace{-.05in} =&&
\hspace{-.3in}
\sum_{i_m} a_{i_1 \dots i_{m-1} i_m i_{m+1} \dots i_M} b_{j_m i_m}, \nonumber\\
\vspace{-.1in}
\ten C = \ten A \times_m \mat B. 
\hspace{+.05in}
\xleftrightarrows[\text{\scriptsize matrixize}][\text{\scriptsize tensorize}]& &
\hspace{-.25in}
\matize C m = \mat B \matize A m.
\nonumber\label{eq:mode-m-product}
\vspace{-.25in}
\end{eqnarray}
\vspace{-.35in}
\end{definition}
\noindent

The $M$-mode SVD (aka. the Tucker decomposition) 
is a ``generalization'' of the conventional matrix (i.e.,
2-mode) SVD which may be written in tensor notation as
\begin{eqnarray}
&\hskip-.3in \mat D&\hskip-.0in= \mat U_1 \mat S \mat U_2\tp \hskip+.225in \Leftrightarrow\hskip+.2in \mat D = \mat S \times_1 \mat U_1 \times_2 \mat U_2.
\end{eqnarray}
The $M$-mode SVD orthogonalizes the $M$ spaces and decomposes the
tensor as the {\it mode-m product}, denoted $\times_m$ 
, of $M$-orthonormal mode matrices, and a core tensor $\ten Z$
\begin{eqnarray}
&\hskip-.3in \ten D&\hskip-.0in={\ten Z} \times_1 {\mat U}_1 \times_2 {\mat U}_2
\dots\times_m {\mat U}_m \dots \times_M {\mat U}_M.
\label{eq:tensor-decomposition}
\end{eqnarray}
\vskip+0.1in

\section{Hierarchical Block Tensor Factorizations of $\ten D$}
\label{sec:global-tensor}
\noindent
Within the tensor mathematical framework, 
a $M$-way array or ``data-tensor'', $\ten D\in\Reals^{I\measure \times I\mode1 \dots \times I\mode c \dots \times I\mode C}$ 
contains a collection of 
 vectorized and centered observations,\footnote{
Reference~\cite[Appendix A]{Vasilescu09} evaluates some of the arguments found in highly cited publications in favor of treating an
image as a matrix (tensor) rather than a vector. While technically speaking, it is not incorrect to treat an image as a matrix, most arguments do not stand up to analytical scrutiny, and it is preferable to vectorize an image and treat it as a single observation rather than a collection of independent column/row observations. 
} 
 $\vec d_{i\mode1\dots i\mode c\dots i\mode C} \in\Reals^{I\measure}$ that are the result of $C$ causal factors. The $c\th$ causal factor ($1\le c\le C$) takes one of $I_c$ values that are indexed by $i_c$, $1\le i_c\le I_c$. 
 An observation that is result of the confluence $C$ causal factors is modeled by a multilinear structural equation with multimode latent variables, $\vec r\mode c$, that represent the causal factors 
 \begin{equation}
 \vec d_{i\mode{1},\dots, i\mode{c},\dots,i\mode{C}} 
 = \ten T \times\mode 1 \vec r\mode 1\T \dots \times \mode c \vec r\mode c\T \dots \times\mode C \vec r\mode C\T +{\bm \epsilon}_{i\mode{1},\dots, i\mode{c},\dots,i\mode{C}},
 \end{equation}
\noindent
where $\ten T=\ten Z\times\measure \mat U\measure$ is the extended core which modulates the interaction between the latent variables, $\vec r\mode c$, that represent the causal factors and $\vec {\bm\epsilon}_{i\mode{1},\dots, i\mode{c},\dots,i\mode{C}} \in \ten N( \vec 0,\mat \Sigma )$ is an
 additive identically and independently distributed (IID) Gaussian noise, Fig.~\ref{fig:tensorfaces}.

\vspace{+.06in}
\subsection{Hierarchical Data Tensor, $\ten D\tmode H$}
\begin{figure*}[!t]
\centering
\centerline{
\begin{tabular}{l}
\hskip +.15in
\includegraphics[width=.51\linewidth]{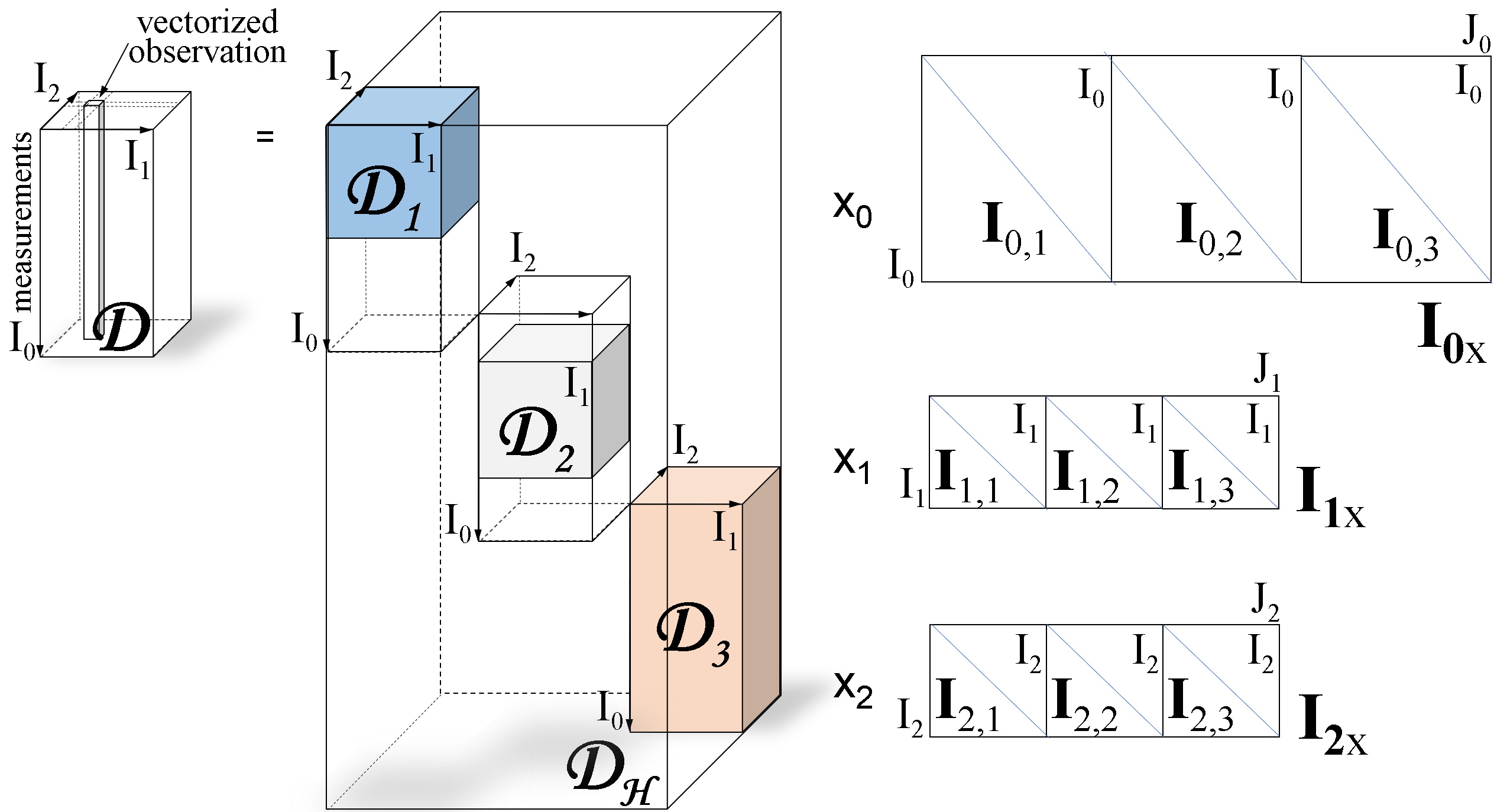}
\\
\end{tabular}
\hfill
\hskip -.11in
\begin{tabular}{r}
\includegraphics[width=.48\linewidth]
{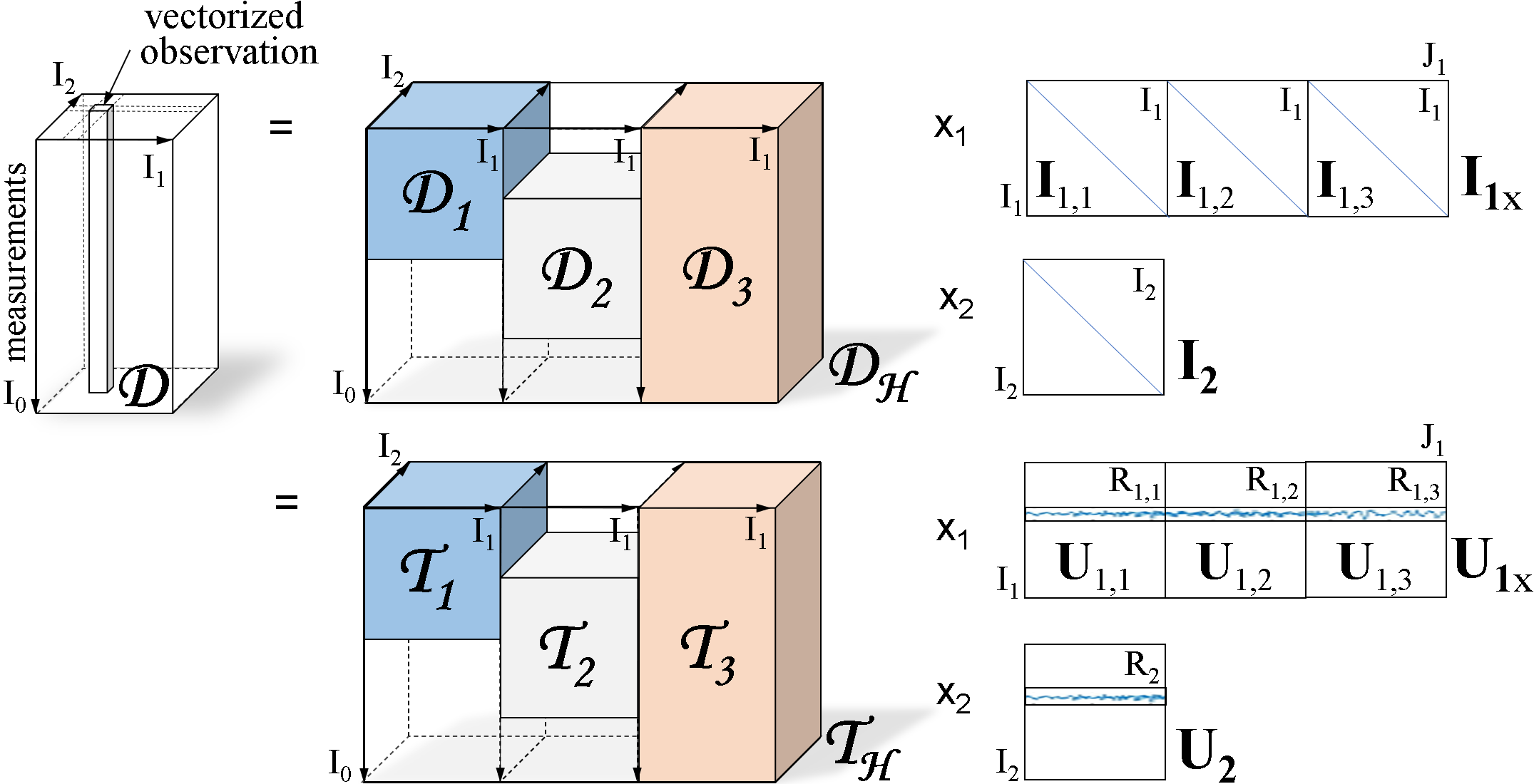}
\hskip+.1in
\hfill
\\
\\
\end{tabular}}
\centerline{\footnotesize \hspace{1.6in}(a)\hfill (b) \hspace{1.6in}}
\vspace{-.005in}
\centerline{
\includegraphics[width=\linewidth]
{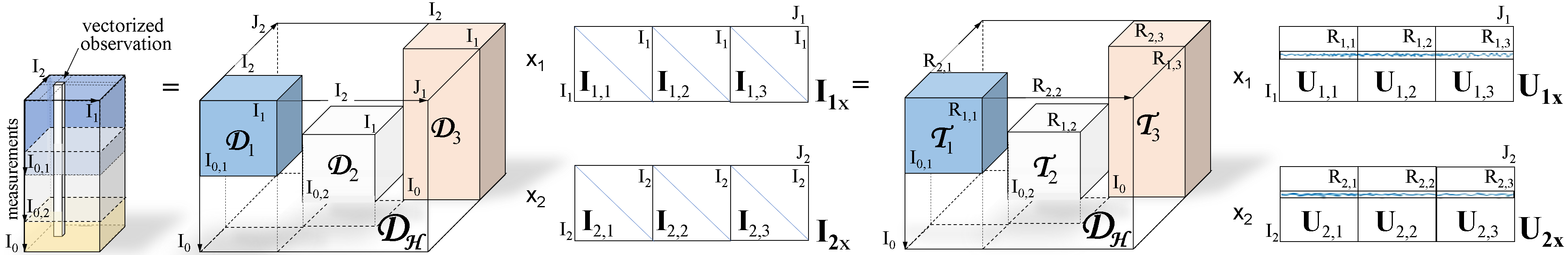}
}
\centerline{\footnotesize \hfill (c) \hfill}

\vspace{+.05in}
\centerline{
\hfill
\includegraphics[height=.165\linewidth]{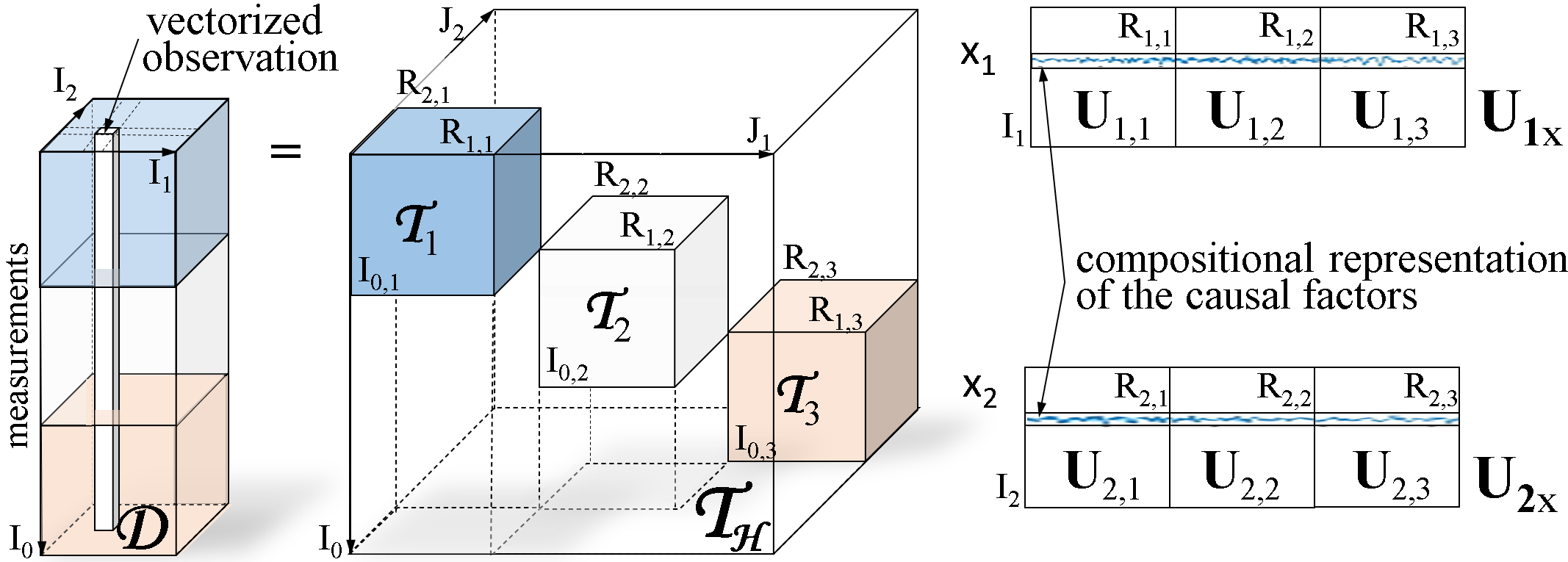}
\hfill
\hskip +.3in
\includegraphics[height=.165\linewidth]{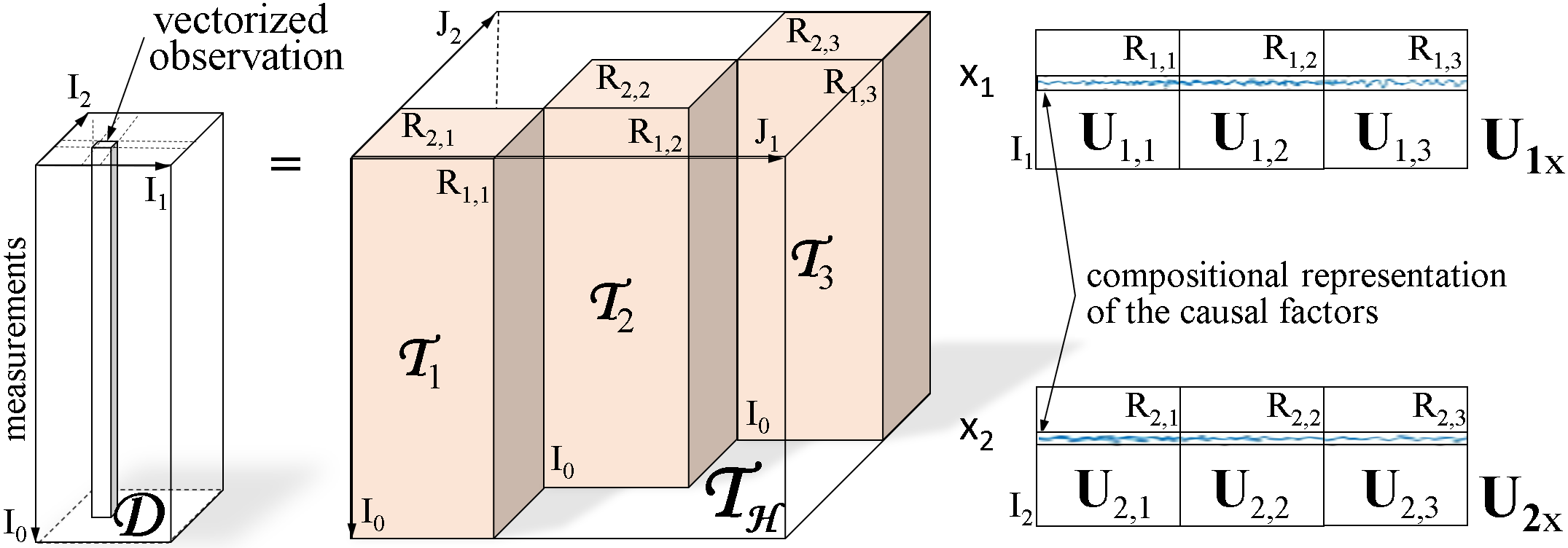}
\hfill
}
\centerline{\footnotesize \hspace{1.6in}(d) \hfill (e)\hspace{1.65in}}
\vspace{-.05in}
\caption[]{
The data tensor, $\ten D$, written in terms of a hierarchical data tensor, $\ten D\tmode H$. (a) When $\ten D\tmode H$ contains the data tensor segments,$\ten D_s$, along its super-diagonal
then $\ten D\tmode H$ has a {\it fully compositional } form, 
and every mode matrix has a compositional  representation. 
(b) A general base case object written in a {\it partially compositional } form with a compositional  representation for only one mode matrix (causal factor). 
(c) A general base case object where all the causal factors have a compositional  representation. The tensor $\ten D\tmode H$ is {\it fully compositional  in the causal factors}.
 (d) A base case object with non-overlapping parts. All the causal factors have a compositional  representations. Multilinear factorizations are block independent. 
(e) Base case object with completely overlapping parts where all the causal factors have compositional  representation. Objects with non-overlapping or completely overlapping parts may also be written using a partially compositional  form analogously to (b).
}
\label{fig:Block-Tucker-Base-Case}
\label{fig:fully_compositional}
\label{fig:independentparts} 
\vspace{-.2in}
\end{figure*}

\vspace{-.025in}
We identify a general base case object and two special cases. 
A base case object may be composed of
(i) two partially overlapping children-parts and parent-whole that has data not contained in any of the children-parts, 
(ii) a set of non-overlapping parts, or
(iii) a set of fully overlapping parts. The tensor representation of an object with fully overlapping parts, Fig.~\ref{fig:Block-Tucker-Base-Case}(e), resembles the rank-$(L,M,N)$ or a rank-$(L,M,\cdot)$ block tensor decomposition~\cite{Delathauwer08b}.\footnote{
The block tensor decomposition~\cite{Delathauwer08b} goal is to find the best fitting $K$ 
fully overlaping tensor blocks that are all multilinearly decomposable into the same multilinear rank-$(R_1,R_2,R_3)$. 
This is analogous to finding the best fitting $K$ rank-$1$ terms (also known as rank-$(1,1,1))$ computed by the CP-algorithm.}

The data wholes and parts are extracted by employing a filter bank 
$\{ \mat H\mode s\in\Reals^{I\measure\times I\measure}\mid\sum_{s=1}^S \mat H\mode s = \mat I,\mbox{ and } 1 \le s \le S \}$ 
where
 a $1D$ ($2D$ or $3D$) convolutional filter, $\vec h\mode s$, 
 is written as a circulant matrix (doubly or triply circulant matrix), $\mat H\mode s$, 
  and $s$ is a segment index. The convolution
may be written as  a matrix-vector multiplication 
or the mode-$m$ product, $\times\mode m$, between a circulant matrix, $\mat H\mode s$ and a vectorized observation. 
For example, if an observation is returned by the capture device as a $2$-way multivariate array, $\mat D\in\Reals^{I\mode{xr} \times I\mode{xc}}$, with $I\mode{xr}$ rows and $I\mode{xc}$ columns, the convolution is written as
\vspace{-.05in}
\begin{eqnarray}
\mat D\mode s =\mat D\ast \vec h_s (x,y) 
\xleftrightarrows[\text{\scriptsize vectorize}][\text{\scriptsize matrixize}]
\vec d\mode s %
=
\mat H\mode s \hspace{+.025in}\vec d
=
\vec d \times\measure \mat H\mode s
\end{eqnarray}
\vspace{-.15in}
\\
\noindent
where the measurement mode is mode $0$. 
In practice, a convolution is efficiently implemented using a DFFT. 
The segment data tensor, $\ten D_s=\ten D \times\measure \mat H\mode s$, is the result of multiplying (convolving) every observation, $\vec d$, with the block circulant matrix (filter), $\mat H\mode s$ ($\vec h\mode s$).  A  filter  $\mat H\mode s$  may be of any type, and have any spatial scope. 
When a filter matrix 
is a block identity matrix, $\mat H\mode{s}=\mat I\mode s$, the filter matrix multiplication with a vectorized observation has the effect of segmenting a portion of the data. 
Measurements associated with perceptual parts may not be tightly packed into a block apriori, as in the case of vectorized images, but chunking is achieved by a trivial permutation.

\vspace{-.075in}
 
 A data tensor is expressed as a recursive hierarchy of wholes and parts by defining and employing a {\it hierarchical data tensor}, $\ten D\tmode H$. 
When a data tensor contains along its super-diagonal the data tensor segments, $\ten D\mode s$, then $\ten D\tmode H$ has a {\it fully compositional } form, and  
all the data tensor modes have a  compositional  representation, 
Fig.~\ref{fig:Block-Tucker-Base-Case}(a).  The data tensor segments, $\ten D_{s}$, may be sparse and represent local parts, or may be full and correspond to a filtered version of a parent-whole, as in the case of a Laplacian pyramid. Mathematically writing $\ten D$ in terms of $\ten D\tmode H$ is expressed with  
\vspace{-.03in}
\begin{eqnarray}
\ten D &=&
\sum_{s=1}^S
\ten D \times\mode 0 \mat H\mode s \label{eq:D_conv}\\
&=&
\ten D\mode {1} \dots + \ten D\mode {s}  \dots + \ten D\mode {S}
\label{eq:D_segment_rep}\\
\hskip -.1in &=& 
\ten D{\tmode H} \times\mode 0 \mat I\mode{Ox}\times\mode 1 \mat I\mode{1x} \dots \times\mode c \mat I\mode{cx} \dots \times\mode C \mat I\mode{Cx}
,\label{eq:DF_rep}
\vspace{-.075in}
\end{eqnarray}
\noindent
where $\mat I\mode{cx}=[ \mat I\mode {c,1} ... \mat I\mode {c,s} ... \mat I\mode {c,S}]\in\Reals^{I_c\times SI_c}$ is a concatenation of $S$ identity matrices, one for each data segment.  
In practice, the measurement mode will not be written in compositional  form, ie. the multipication with  $\mat I\measure{}\pixels$ would have been carried out.
The resulting $\ten D\tmode H$ is {\it fully compositional  in the causal factors}, where  every causal factor has a compositional representation rather than every mode Fig.~\ref{fig:Block-Tucker-Base-Case}(c).   Articulated-objects have parts with their own extrinsic causal factors 
and benefit from a compositional representation of every causal factor.
A non-articulated object where the wholes, and parts share the same extrinsic causal factor representations (same illumination/viewing conditions)
benefit from being written in terms of {\it a partially compositional  data tensor}, 
where the intrinsic causal factor has a  compositional  form, the intrinsic object representation, Fig.~\ref{fig:Block-Tucker-Base-Case}(b). Thus, the $\ten D\tmode H$ is multiplied through by all the $\mat I\mode{cx}$ except one. Each multiplied $\mat I\mode{cx}$ is replaced by a single place holder identity matrix in the model.

The three different ways of rewriting $\ten D$ in terms of a hierarchy of wholes and
parts, eq.~\ref{eq:D_conv}-\ref{eq:DF_rep}, 
results in three mathematically equivalent representations~\footnote{\label{note:equivalence}$
{\mbox{Equivalent representations}}$ can be transformed into one another by post-multiplying mode matrices with 
nonsingular matrices, $\mat G\mode c$,\\ $\ten D= ({\ten Z}_{\ten H}\times\mode 0 \mat G\inv\mode 0 \dots \times\mode c \mat G\inv\mode c \dots \times\mode C \mat G\inv\mode C) \times\mode 1  \mat I\mode{1x}{\mat U\mode 1}{}_{\ten H}\mat G\mode 1 \dots \times\mode c  \mat I\mode{cx}{\mat U\mode c}_{\ten H} \mat G\mode c \dots \times\mode C  \mat I\mode{Cx}{\mat U\mode C}_{\ten H} \mat G\mode C$.}  
based on 
factorizing 
$\ten D$, $\ten D\mode s$ and $\ten D\tmode H$:
\\
\vspace{-.15in}
\begin{eqnarray}
\hskip-.8in\ten D 
&\hskip-.1in=&\hskip-.1in
\sum_{s=1}^S
\underbrace{(\ten Z\times\measure \mat U\measure \times \mode 1 \mat U\mode 1 \dots \times\mode c \mat U\mode c \dots \times\mode C \mat U\mode C )}_{\mbox{
$\ten D$}}\times\mode 0  
{\mat H\mode s} \label{eq:D_decomp}\\
&\hskip-.1in=&\hskip-.1in
\sum_{s=1}^S
\underbrace{(\ten Z\mode s\times\measure \mat U\measure{}\mode{,s} \times \mode 1 \mat U\mode {1,s} \dots \times\mode c \mat U\mode {c,s} \dots \times\mode C \mat U\mode {C,s} )}_{\mbox{
$\ten D\mode s$}}
\label{eq:Ds_decomp}\\
&\hskip-.1in=&\hskip-.1in
\underbrace{(\ten Z\tmode H\times\measure \mat U\measure{}\tmode H \times \mode 1 \mat U\mode {1}{}\tmode H \dots \times\mode c \mat U\mode {c}{}\tmode H \dots \times\mode C \mat U\mode {C}{}\tmode H)}_{\mbox{
$\ten D\tmode H$}}\nonumber\\
&&\hskip+.2in\times\mode 0 \mat I\mode{0x}\times\mode 1 \mat I\mode{1x} \dots \times\mode c \mat I\mode{cx} \dots \times\mode C \mat I\mode{Cx}
\label{eq:DH_decomp}
\vspace{-.1in}
\end{eqnarray}
\vspace{-.15in}
\\
Despite the prior mathematical equivalence, equations~\ref{eq:D_conv},\ref{eq:D_decomp},
and equations~\ref{eq:D_segment_rep},\ref{eq:DH_decomp} 
are not flexible enough to explicitly indicate if the parts are organized in a partially compositional  form, or a fully compositional  form. 

The expression of $\ten D$ in terms of a hierarchical data tensor is a mathematical conceptual device,
 that enables a unified mathematical model of wholes and parts
that can be expressed completely as a mode-m product (tensor-matrix multiplication) and whose factorization can be optimized in a principled manner. 

Dimensionality reduction of the compositional  representation 
is performed by optimizing 
\begin{eqnarray}
&\hskip-.4in e& 
    \hskip -.25in = \hskip -.125in
    =
    \|\ten D \hskip-.025in- \hskip-.025in
    (
\hskip+.025in \bar{\ten Z}\tmode H  \times\measure \bar{\mat U}\mode 0{}\tmode H 
\times\mode 1 \bar{\mat U}\mode {1}{}\tmode H ...  \times\mode C \bar{\mat U}\mode{C}{}\tmode H)
\times\mode 0 \mat I\mode{0x} 
... \times\mode C \mat I\mode {Cx}
\|^2
\label{eq:loss_fnc}
\nonumber \\
& &\hskip+.2in 
    + \sum_{c=0}^C \lambda\mode {c}\|{{\bar{\mat U}\mode c}{}\tp\tmode H}{\bar{\mat U}\mode c}{}\tmode H -\mat I\|^2 
\label{eq:loss_fnc_DH}
\vspace{-.25in}
\end{eqnarray}

\noindent
\vskip-.25in
where $\bar{\mat U}\mode{c}{}\tmode{H}$ is the
composite representation of the $c^{\mbox{\small th}}$ mode, and $\ten Z\tmode H$ governs the interaction between causal factors.
Our optimization may be initialized by setting $\ten Z\tmode H$ and $\mat U\mode c{}\tmode H$ to the M-mode SVD of $\ten D\tmode H$,\footnote{Note that eq.(\ref{eq:loss_fnc_DH}) does not reduce to a multilinear subspace decomposition of $\ten D\tmode H$ since $\ten D \times\mode 0 \mat I\mode{Ox}\pinv{} \times\mode 1 \mat I\mode{1x}\pinv{} \dots \times\mode c \mat I\mode{cx}\pinv{} \dots \times\mode C \mat I\mode{Cx}\pinv{} \neq \ten D{\tmode H}$.}
~\footnote{For computational efficiency, we may perform M-mode SVD on each data tensor segment  $\ten D\mode s$ and concatenate terms along the diagonal of $\ten Z\tmode H$ and $\mat U\mode c{}\tmode H$. 
 } 
 and performing dimensionality reduction through truncation, where $\bar{\mat U}\mode c{}\tmode H \in 
 \Reals^{SI_c \times \bar{J}_c}$, 
 $\bar{\ten Z}\tmode H \in \Reals^{\bar J_ 0 \dots \times \bar J_c \dots \times \bar J_C}$ and $\bar J\mode c \le SI\mode c$.
\noindent


\noindent
\subsection{Derivation}
For notational simplicity, 
we 
re-write the loss function as,
%
\begin{eqnarray}
    e 
    \hskip -.1in &\assign& \hskip -.1in
    \|\ten D - 
    \tilde{\ten Z}{\tmode H} \times\mode 0 \tilde{\mat U}\mode {0x}
...%
\times\mode c \tilde{\mat U}\mode {cx}
...
\times\mode C \tilde{\mat U}\mode {Cx}
    \|^2 \nonumber\\
    \hskip -.1in & & \hskip -.1in 
    + \sum_{c=0}^C\sum_{s=1}^S \lambda\mode {c,s}\|\tilde{\mat U}\mode {c,s}\tp \tilde{\mat U}\mode {c,s}-\mat I\| 
    \label{eq:loss_fnc}
    \vspace{-.4in}
\end{eqnarray}
\vspace{-.05in}

\noindent
where $\tilde{\mat U}\mode {cx} = \mat I\mode{cx}\bar{\mat U}\mode c{}\tmode H \tilde{\mat G}\mode c=[\tilde{\mat U}\mode{c,1}| \dots |\tilde{\mat U}\mode{c,s}| \dots |\tilde{\mat U}\mode{c,S}]$ and $\tilde{\mat G}\mode c\in\Reals^{\bar J_c \times SI_c}$ is permutation matrix that groups the columns of $\mat U\mode c{}\tmode H$ based on the segment, $s$, to which they belong, and the inverse permutation matrices have been multiplied${}^{\ref{note:equivalence}}$ into $\tilde{\ten Z}\tmode H$ 
resulting into 
a core that has  also been grouped based on 
segments and sorted based on variance.  
\noindent
The data tensor, $\ten D$, may be expressed in matrix form as in eq.~\ref{eq:Dapprox_matrix1} and reduces to 
the more efficiently block structure as in eq.~\ref{eq:Dapprox_matrix2}
\begin{eqnarray}
&\hskip-.25in {\ten D}
&\hskip -.15in = 
{\ten Z}\tmode H\times\mode 0 {\mat U}\mode{0x} \times\mode 1 {\mat U}\mode{1x}\dots \times\mode c {\mat U}\mode{cx}\dots \times\mode C {\mat U}\mode{Cx}
\\
&\hskip-.15in 
{\mat D}\mode{[c]} &\hskip-.15in 
={\mat U}\mode {cx} {\mat Z}\tmode{H}{}\mode{[c]}
\left({\mat U}\mode {Cx} \otimes \dots \otimes {\mat U}\mode{(c+1)x} \otimes {\mat U}\mode{(c-1)x} \otimes \dots \otimes {\mat U}\mode{0x}\right)\tp \hskip-.305in\label{eq:Dapprox_matrix1}\\
&\hskip-.15in
=& \hskip-.15in
\left[\hskip+.025in {\mat U}\mode{c,1} \dots {\mat U} \mode{c,s} \dots {\mat U} \mode{c,S}\hskip+.025in\right]\hskip-.305in\label{eq:Dapprox_matrix2}
\\
&&\hskip-.4in
\left[\hskip -.075in
\begin{array}{ccccc}
{\mat Z}\mode{0[c]}\pinv{} &\hskip-.1in\mat 0& \hskip -.1in&  \hskip -.1in\cdots&\hskip-.1in\mat 0 \\
\mat 0 &\hskip-.1in\ddots&  \hskip -.1in\mat 0&  \hskip -.1in &\hskip-.1in\vdots  \\
\vdots  &\hskip-.1in\mat 0 & \hskip -.1in{\mat Z}\mode{s[c]}\pinv{}& \hskip -.1in \mat 0 & \hskip-.1in \\
 &\hskip-.1in &  \hskip -.1in\mat 0 &  \hskip -.1in\ddots&  \hskip-.1in\mat 0\\
\mat 0 & \hskip-.1in\cdots & \hskip -.1in &  \hskip -.1in\mat 0& \hskip-.1in{\mat Z}\mode{S[c]}\pinv{}
\end{array}\hskip -.075in\right]
\hspace{-.045in}
\underbrace{\left[\hskip -.075in
\begin{array}{c}
\left({\mat U}\mode {C,1} 
...
\otimes {\mat U}\mode {(c+1),1} \otimes {\mat U}\mode {(c-1),1} 
...
\otimes {\mat U}\mode {0,1}\right)\tp\\
\vdots\\
\left({\mat U}\mode{C,s} 
...
\otimes {\mat U}\mode{(c+1),s} \otimes {\mat U}\mode{(c-1),s} 
...
\otimes {\mat U}\mode{0,s}\right)\tp\\
\vdots \\
\left({\mat U}\mode{C,S} 
...
\otimes {\mat U}\mode{(c+1),S} \otimes {\mat U}\mode{(c-1),S} 
...
\otimes {\mat U}\mode{0,S}\right)\tp\\
\end{array} \hskip -.075in \right]}
\hskip-.075in\label{eq:independent_efficient}
\nonumber\\
&&
\hskip+1in{
\left({\mat U}\mode {Cx} 
\dots
\odot {\mat U}\mode {(c+1)x} \odot {\mat U}\mode {(c-1)x} 
\dots 
\odot {\mat U}\mode {0x}\right)\tp}
\nonumber\\
&\hskip-.1in = 
{\mat U}\mode {cx} \mat W\mode c\tp,\hskip-.35in
\vspace{-.5in}
\end{eqnarray}

\vspace{-.1in}
where $\otimes$ is the Kronecker product\footnote{$
{\mbox{
The Kronecker product}}$ of $\mat U \in \Reals^{I\times J}$ and 
$\mat V \in \Reals^{K\times L}$ is the $IK \times JL$ 
matrix defined as $[\mat{U} 
\otimes 
\mat{V}]_{ik,jl} = u_{ij} v_{kl}$.
}, 
and 
$\odot$ 
is the 
block-matrix Kahtri-Rao product.\footnote{The Khatri-Rao product of 
$\left[\mat U\mode1 \dots \mat U\mode n \dots \mat U\mode N\right]
\odot 
\left[\mat V\mode1 \dots \mat V\mode n \dots \mat V\mode N\right]$ 
with $\mat U\mode l \in \Reals^{I\times N\mode l}$ and $\mat V\mode l \in 
\Reals^{K\times N\mode l}$ 
is a {\it block-matrix Kronecker product}; therefore, it 
can be
expressed as $\mat U \odot \mat V = [(\mat U\mode{1}\otimes \mat V\mode{1}) \dots (\mat U\mode{(l)}\otimes \mat V\mode{(l)}) \dots (\mat
U\mode{(L)}\otimes \mat V\mode{(L)})]$.
} 


The matrixized block diagonal form of ${\ten Z}\tmode H$ in eq.~\ref{eq:Dapprox_matrix2} becomes evident when employing our modified data centric matrixizing operator  based on the defintion~\ref{def:matrixizing}, where the initial mode is the measurement mode.

The hierarchical block multilinear factorization, the $M$-mode Block SVD algorithm computes the mode matrix, ${\mat U}\mode{cx}$, by computing the 
minimum of $e=\|\ten D
- \tilde{\ten Z}\tmode H \times\mode 0 \tilde{\mat U}\mode {0x} \dots \times\mode C \tilde{\mat U}\mode {Cx}\|^2$ by cycling
through the modes, solving for $\tilde{\mat U}\mode {cx}$ in the equation $\partial
e/\partial \mat U\mode {cx}=0$ while holding the core tensor $\ten Z\tmode H$ and all the other mode matrices
constant, and repeating until convergence. Note that
\begin{eqnarray}
\hskip-.055in
{\partial e\over\partial \mat U\mode {cx}}\hskip-.025in&=&\hskip-.025in{\partial\over\partial \mat U\mode {cx}}
\|{\matize D c} - \mat U\mode {cx}\mat W\mode c\tp\|^2 
=-{\matize D c}
\mat W\mode c + \mat U\mode {cx} \mat W\mode c\tp \mat W\mode c.
\nonumber
\end{eqnarray}

\noindent
Thus, $\partial e/\partial \mat U\mode {c}{}\pixels=0$ implies that

\hskip-3in
\vspace{-.2in}
\begin{eqnarray}
&\hspace{-.1in}\mat U\mode {cx}
&= 
{\matize D c} \mat W\mode c {\left(\mat W\mode c\tp \mat W\mode c\right)\hskip-.025in}^{-1} 
\hskip-.085in = \hskip-.02in
{\matize D c} \mat W\mode c\tp{}\pinv{}
\hskip-.5in
\label{eq:U_general1}\nonumber\\ 
\hskip-.15in&
\hskip-.15in
=
&
\hskip-.2in 
{\matize D c} \hskip-.01in{\left( \hskip-.025in\mat Z\tmode H{}\mode {[c]} \hskip-.025in\left({\mat U}\mode {Cx} \hskip-.025in\otimes\hskip-.025in 
...
\hskip-.025in{\mat U}\mode {(c+1)x} \hskip-.025in\otimes \hskip-.025in{\mat U}\mode {(c-1)x} \hskip-.025in\otimes \hskip-.025in
...
\hskip-.025in{\mat U}\mode {0x}\right)\tp \hskip-.02in\right)\hskip-.05in}\pinv{}
\hskip-.5in\label{eq:U_general2}\nonumber 
\\ 
&
\hskip-.15in
= &
\hskip-.2in
{\matize D c}\hskip-.025in
\left({\mat U}\mode {Cx} 
\hskip-.05in\odot
...
{\mat U}\mode {(c+1)x} \hskip-.025in\odot \hskip-.025in{\mat U}\mode {(c-1)x} 
\hskip-.025in\odot
...
{\mat U}\mode {0x}\right)\hskip-.05in\tp{}\pinv{}
\hskip-.025in\left[\hskip -.075in
\begin{array}{ccccc}
{\mat Z}\mode{0[c]}\pinv{} &\hskip-.1in\mat 0& \hskip -.1in&  \hskip -.1in\cdots&\hskip-.1in\mat 0 \\
\mat 0 &\hskip-.1in\ddots&  \hskip -.1in\mat 0&  \hskip -.1in &\hskip-.1in\vdots  \\
\vdots  &\hskip-.1in\mat 0 & \hskip -.1in{\mat Z}\mode{s[c]}\pinv{}& \hskip -.1in \mat 0 & \hskip-.1in \\
 &\hskip-.1in &  \hskip -.1in\mat 0 &  \hskip -.1in\ddots&  \hskip-.1in\mat 0\\
\mat 0 & \hskip-.1in\cdots & \hskip -.1in &  \hskip -.1in\mat 0& \hskip-.1in{\mat Z}\mode{S[c]}\pinv{}
\end{array}\hskip -.075in\right]\hspace{+.4in}
\nonumber
\label{eq:U_general}\hskip-.2in
\end{eqnarray}
\noindent
whose $\mat U\mode{c,s}$ sub-matrices are then subject to 
orthonormality constraints.   \\
\begin{figure}[!t]%
\vspace{-.025in}
    \centering
    \includegraphics[width=1\linewidth]{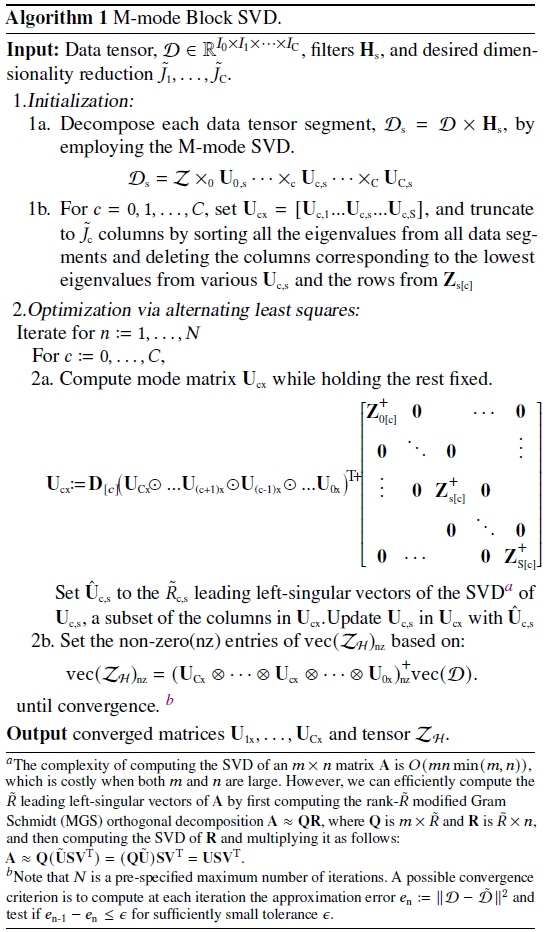}
    \label{fig:Block_M_mode_SVD}
   \vspace{-.175in}
   \vspace{-.2in}
\end{figure}
Solving for the optimal core tensor, $\ten Z\tmode H$, the 
data tensor, $\ten D$, approximation is expressed in 
vector form as,
\vspace{-.025in}
\begin{eqnarray}
e=\|\mathrm{vec}({\ten D}) - (\tilde{\mat U}\mode{Cx}
\otimes \dots \otimes\tilde{\mat U}\mode{cx}\otimes 
\dots\otimes \tilde{\mat U}\mode{0x})\mathrm{vec}(\tilde{\ten Z}\tmode H)\|.
\end{eqnarray}
\noindent
Solve for the non-zero(nz) terms of $\ten Z\tmode {H}$ in the equation $\partial e/\partial ({\small\ten Z\tmode H}) =0$,
by removing the corresponding zero columns of the first matrix on right side of the equation below, performing the pseudo-inverse, 
and setting
\begin{eqnarray}
\mathrm{vec}(\ten Z\tmode H)\mode{nz}=(\mat U\mode{Cx} 
\otimes \dots \otimes\mat U\mode{cx}\otimes 
\dots\otimes \mat U\mode{0x})\pinv{}\mode{nz}\mathrm{vec}(\ten D).
\label{eq:tensor_model}
\end{eqnarray}

\vskip-.1in
\noindent
Repeat all steps until convergence.This optimization is the basis of the 
$M$-mode Block SVD Algorithm 1.
\vspace{\belowdisplayskip}\hfill
\vskip-.2in
\rule{\dimexpr(.5\textwidth-0.5\columnsep-1pt)}{0.4pt}%
\rule{0.4pt}{6pt}
\newpage
\newpage
When the data tensor is a collection of observations made up of non-overlapping parts, Fig.~\ref{fig:Block-Tucker-Base-Case}d, the data tensor decomposition reduces to the concatenation of an M-mode SVD of individual parts and when the data tensor is a collection of overlapping parts that have the same multilinear-rank reduction, Fig.\ref{fig:Block-Tucker-Base-Case}e, see\cite{Vasilescu19} for additional specific optimizations.

\vspace{+.05in}
\section{
Representing Levels of Abstraction Bottom-up}
An incremental hierarchical block multilinear factorization that represents levels of abstractions bottom-up is developed analogously to the incremental SVD for matrices~\cite{Brand02}.  The precomputed multilinear factorizations of the children parts are employed to determine the parent whole multilinear factorization.  The derived algorithm may also be employed to update the overall model when the data becomes available sequentially~\cite{Li07}. 
We first address the computation of the mode matrices and the extended core of the parent whole when the children parts are non-overlapping. Next, we consider the overlapping children case, and the case where the parent-wholes and children-parts contain differently filtered data. 
\vspace{-.05in}
\noindent
{\bf Computing parent causal mode matrices, $\mat U\mode{c,w}$:}  
Note that the parent whole, $\ten D_w$, is a concatenation of 
the data contained in its $K'$ children segments that are part of the hierarchy, $\ten D_k$, where $1\le k\le K'$.  New data that is not contained by any of the children is denoted as the $K=K'+1$ child,$\ten D\mode K$, eq.~\ref{eq:iD_of_parts} .
We initialize the hierarchical block multilinear factorization by performing an $M$-mode SVD on each leaf. 


The $c\th$ mode matrix, $\mat U_{c,w}$ of the $w$ parent whole, $\ten D_w$, is the left singular matrix of
\vspace{-.25in}
\begin{equation}
\hspace{+1.25in}
    \left[\hskip-.025in
    \begin{array}{ccccc}
    \mat U\mode{c,1}\mat \Sigma\mode{c,1} & ... & \mat U\mode{c,k}\mat \Sigma\mode{c,k} & ... & \mat U\mode{c,K}\mat \Sigma\mode{c,K}
 \end{array}\hskip-.025in\right]\nonumber
\vspace{-.025in}
\end{equation} 
which is based on the following derivation, that writes SVD of the flattened parent whole in terms of the SVDs of its flattened children parts, followed by a collection terms such that $\mat V\mode{c,all}$ is a block diagonal matrix of $\mat V\mode{c,k}$: 
\begin{eqnarray}
\hspace{-.0in}
\matize{D\mode{w}} c 
\hskip-.05in
&=&
\hskip-.025in
\left[\hskip-.025in\begin{array}{ccccc}
{\mat D\mode{1}}{}\mode{[c]} & \hskip-.025in \cdots \hskip-.025in
&{\mat D\mode{k}}{}\mode{[c]} & \hskip-.025in \cdots \hskip-.025in
& {\mat D\mode{K}}{}\mode{[c]} 
\end{array}\hskip-.025in\right]\hskip-.25in\label{eq:iD_of_parts}\\
&=&
\hskip-.025in
    \left[\hskip-.025in
    \begin{array}{cccccc}
    \mat U\mode{c,1}\mat \Sigma\mode{c,1}\hskip-.025in\mat V\mode{c,1}\tp
    &\cdots
    &\mat U\mode{c,k}\mat \Sigma\mode{c,k}\hskip-.025in\mat V\mode{c,k}\tp
    &\cdots 
    &\mat U\mode{c,K}\mat \Sigma\mode{c,K}\hskip-.025in\mat V\mode{c,K}\tp 
    \end{array}\hskip-.05in
    \right]\nonumber\\
    &=& 
    \hskip-.025in
    \underbrace{\left[\hskip-.05in
    \begin{array}{ccccc}
    \mat U\mode{c,1}\mat \Sigma\mode{c,1} & \hskip-.0in\dots & \hskip-.0in\mat U\mode{c,k}\mat \Sigma\mode{c,k} & \hskip-.0in\dots & \hskip-.0in\mat U\mode{c,K}\mat \Sigma\mode{c,K}
    \end{array}\hskip-.05in\right]}_{\mbox{QR + SVD of R}} \mat V\mode{c,all}\tp\hskip-.35in\label{eq:D_parent_QR_SVD}\\
    &=&
\mat U\mode{c,w}\mat \Sigma\mode{c,w}
   \underbrace{ \left[
    \begin{array}{ccccc}
\mat V\mode{c,w1}\tp & \hskip-.0in\dots & \hskip-.0in\mat V\mode{c,wk}\tp & \hskip-.0in\dots & \hskip-.05in\mat V\mode{c,wK}\tp
    \end{array}\right] \mat V\mode{c,all}\tp}_{\mat V\mode{c,w}\tp},   
\hskip-.25in\label{eq:U_parent_inc}
\end{eqnarray}

\vspace{-.1in}
{\bf Computing the parent extended core, $\ten T\mode w$:} Computation of the extended core associated with the parent whole, $\ten T\mode w$, is performed by considering the following derivation

\vspace{-.125in}
\begin{eqnarray}
\hskip+3in\label{eq:D_parent_fake2}
\end{eqnarray}
\vskip-.9in
\begin{strip}
\begin{align*}
\vspace{-.145in}
\hskip-.0in{\mat D\mode{w}}&{}\mode{[c]}
=%
\hskip .05in 
\left[\hskip-.05in\begin{array}{ccccc}
{\mat D\mode{1}}{}\mode{[c]} & \dots 
&{\mat D\mode{k}}{}\mode{[c]} & \dots 
& {\mat D\mode{K}}{}\mode{[c]}
\end{array}\hskip-.05in\right]\nonumber\\
=&
\hskip -.025in
\left[\hskip-.05in\begin{array}{cccccc}
\mat U_{c,1}\mat \Sigma_{c,1}\hat{\mat T}\mode{1[c]}
(\mat U\mode{1,1}\otimes ...
\mat U\mode{c-1,1}\otimes\mat U\mode{c+1,1} \otimes 
...
\mat U\mode{C,1}\
)\tp
&\dotsm&
\mat U_{c,k}\mat \Sigma_{c,k}\hat{\mat T}\mode{k[c]}
(\mat U\mode{c-1,0}\otimes
...
\mat U\mode{c-1,k}\otimes\mat U\mode{c+1,k} \otimes 
...
\mat U\mode{C,k}
)\tp
&\dotsm& 
\end{array}\hskip-.075in\right]
\\
=&
\hskip -.025in 
\underbrace{[\hskip-.05in
\begin{array}{ccc}
\mat U_{c,1}\mat \Sigma_{c,1} 
&
\hskip-.1in 
...
\hskip+.05in 
\mat U_{c,k} \mat \Sigma_{c,k}
\hskip+.05in 
...
& \hskip-.1in  \mat U_{c,K} \mat \Sigma_{c,K}
\end{array}
\hskip-.075in]}_{\mbox{QR + SVD of R}} 
\hskip-.05in\left[\hskip -.075in
\begin{array}{ccc}
{\hat{\mat T}}\mode{1}{}\mode{[c]} 
(\mat U\mode{1,1} \otimes 
...
\mat U\mode{c-1,1}\otimes\mat U\mode{c+1,1}\otimes
...
\mat U\mode{C,1})\tp 
&\mat 0 & {\dots}_{\vdots}\\
\hskip-.2in\mat 0  &\ddots & \hskip+.2in\mat 0 \\
\vdots\dots & \mat 0 
& {\hat{\mat T}}\mode{K}{}\mode{[c]} 
(\mat U\mode{1,K}\otimes 
...
\mat U\mode{c-1,K}\otimes\mat U\mode{c+1,K} \otimes
...
\mat U\mode{C,K})\tp \\
\end{array}\hskip-.075in\right]\hskip+.2in\\    
=&
\mat U\mode{c,w} \mat \Sigma\mode{c,w} 
    \underbrace{
    \left[\hskip-.075in
    \begin{array}{ccccc}
\mat V\mode{c,1}\tp &\dots & \mat V\mode{c,k}\tp & \dots & \mat V\mode{c,K}\tp
\end{array}\hskip-.075in\right]}_{\mat V\mode c\tp}
\hskip-.025in
\left[\hskip -.075in
\begin{array}{ccc}
{\hat{\mat T}}\mode{1}{}\mode{[c]} 
(\mat U\mode{1,1}\otimes
...
\mat U\mode{c-1,1}\otimes\mat U\mode{c+1,1}\otimes 
...
\mat U\mode{C,1})\tp 
& \mat 0 & \dots{   }_{\vdots}\\
\hskip-.2in\mat 0 & \ddots & \hskip+.2in\mat 0\\
\vdots \dots& \mat 0 & {\hat{\mat T}}\mode{K}{}\mode{[c]} 
(\mat U\mode{1,K}\otimes 
...
\mat U\mode{c-1,K}\otimes\mat U\mode{c+1,K}
\otimes
...
\mat U\mode{C,K})\tp \\
\end{array}
\hskip-.075in\right].\hskip-.4in\nonumber\label{eq:core_parent} 
%
\end{align*}
\vspace{-.1in}
\end{strip}

\newpage
\noindent
where ${\hat{\mat T}}\mode{k[c]}=\mat \Sigma\mode{c,k}\inv{\mat T}\mode{k[c]}$. Let $\hat{\ten T}\mode{k}={\ten T}\mode{k}\times\mode 1 \mat \Sigma\mode{1,k}\inv \dots \times\mode c \mat \Sigma\mode{c,k}\inv \dots\times\mode C \mat \Sigma\mode{C,k}\inv $ is the normalized extended core of the $k^{\mbox{th}}$ child, and $\hat{\ten T}\mode {Kall}$ 
contains along the diagonal the children normalized extended cores $\hat{\ten T}\mode k$. Thus, the extended core of the parent whole is
\vskip-.25in
\begin{equation}
\ten T\mode w=\hat{\ten T}\mode {Kall} \times\mode 1 \mat \Sigma\mode{1,w} \mat V\mode{1}\tp \dots \times\mode c \mat \Sigma\mode{c,w} \mat V\mode{c}\tp \dots \times\mode C \mat \Sigma\mode{C,w} \mat V\mode{C}\tp.
\end{equation}
\noindent

\vskip-.1in
{\bf Overlapping children:} This case may be reduced to the non-overlapping case by introducing another level in the hierarchy. Overlapping children are now treated as parents with one non-overlaping child sub-part and child sub-parts that correspond to every possible combination of overlaps that are shared by siblings. The original parent whole representation is computed in terms of the grandchildren representations.\\
{\bf Parent-whole and children-parts with differently filtered data:} This is the case when a parent-whole and the children parts contain differently filtered information, as in the case when a parent-whole and the children parts sample information from different layers of a Laplacian pyramid. This case
may be reduced to a non-overlapping case by writing the filters as the product between  a segmentation filter, $\mat S$, \ie an identity matrix with limited spatial scope, and general filter that post multiplies the segmentation filter, $\mat H\mode s= \mat F\mode s \mat S\mode s$ and $\ten D\mode s=(\ten D\times\measure \mat S\mode s)\times\measure \mat F\mode  s$. The general filters, $\mat F\mode  s$ may be applied after the cores are computed. 

\begin{figure*}[t]
\vskip+.05in
\centering
    \includegraphics[width=.875\textwidth]{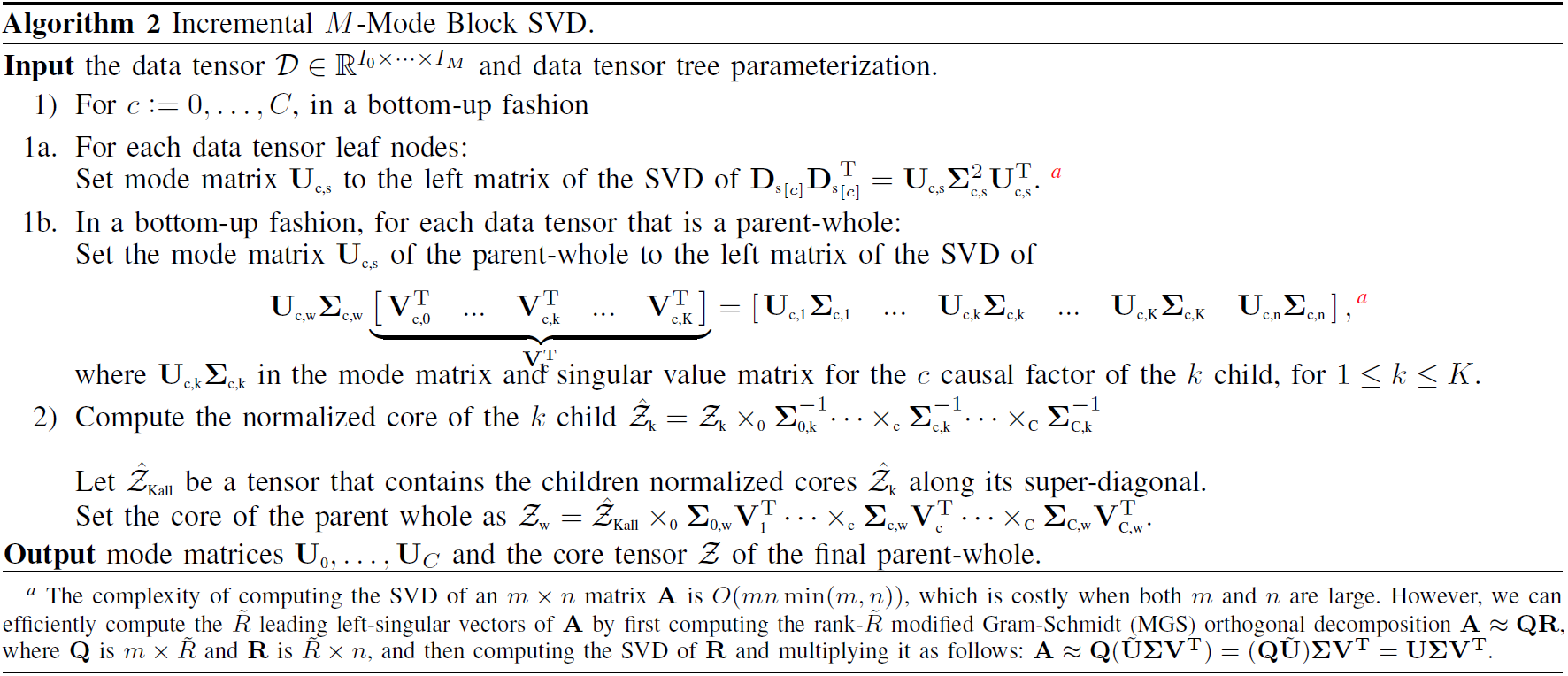}
    \label{fig:incremental_block_m_mode_SVD}
    \vspace{-.245in}
\end{figure*}

\vspace{-.05in}
{\bf Computational Cost Analysis:}
Let an $M$-order data tensor, $\ten D\in \Reals^{I\mode 0 \times I\mode 1\dots \times I\mode c \times \dots \times I\mode C}$, where $M=C+1$, be recursively 
subdivided into $K=2^M$ children of the same order, but with each mode half in size. There are a total of $log_{K}{N}+1$ levels, where $N=\prod_{i=0}^C I_i$. Recursive subdivision results in $S=N\log_{2^M}{N}+1$ segments.
 The total computational cost is the amortized  M-mode SVD cost per data tensor segment, $T$, times the number of segments, $\mathcal{O}(T N \log_{K}N)$. Since siblings at each level can be computed independently, on a distributed system the cost is $\mathcal{O}(Tlog_K N)$. 
\\
\vspace{-.1in}
\section{CausalX Experiments}
\vspace{-.05in} 
CausalX visual recognition system computes a set of causal explanations based on a counterfactual causal model that takes advantage of the assets of  multilinear (tensor) algebra. 
The $M$-mode Block SVD and the Incremental $M$-mode Block SVD algorithms estimate the model parameters.  In the context of face image verification,
we compute a compositional hierarchical person representation~\cite{Vasilescu19}. Our system is trained on a set of observations that 
are the result of combinatorially manipulating the scene structure, the viewing and illumination conditions. We rendered in Maya images of 100 people from 15 different viewpoints with 15 different illuminations. The collection of vectorized images with $10,414$ pixels is organized in a data tensor,
$\ten D\in \Reals^{10,414 \times 15\times 15\times 100}$. 
The counterfactual model is estimated 
by employing $\ten D\tmode H$, a hierarchical tensor of part-based Laplacian pyramids. 
We report encouraging face verification results on two test data sets – the Freiburg, and the Labeled Faces in the Wild (LFW) datasets. We have currently achieved verification rates just shy of $80\%$ on LFW~\cite{Vasilescu19}, 
 by employing 
 less than one percent ($1\%$) of the total images employed by DeepFace~\cite{Taigman14}. 
 When data is limited, convolutional neural networks (CNNs) do not convergence or generalize. More importantly, CNNs are predictive rather than causal models.
\vspace{-.15in}
\section*{Conclusion}
\vspace{-.05in}
This paper deepens the 
definition 
of causality in a multilinear (tensor) framework by addressing 
the distinctions 
between intrinsic versus extrinsic causality, and local versus global causality. 
It proposes a unified multilinear model of wholes and parts that reconceptualizes a data tensor in terms of a {\it hierarchical data tensor}. 
Our hierarchical data tensor is a mathematical instantiation of a tree data structure that enables a single elegant model of wholes and parts and 
allows for different tree parameterizations for the intrinsic versus extrinsic causal factors. The derived tensor factorization is a hierarchical block multilinear factorization that disentangles the causal structure of data formation.  Given computational efficiency considerations, we present an incremental computational alternative that employs the part representations from the lower levels of abstraction to compute the parent whole representations from the higher levels of abstraction in an iterative bottom-up way. This computational approach may be employed to update causal representations in scenarios when data is available incrementally. 
 The resulting object representation is a combinatorial choice of part representations, that renders object recognition robust to occlusion and reduces large training data requirements. We have demonstrated our work in the context of face verification by extending the TensorFaces method with promising results. TensorFaces is a component of CausalX, a counterfactual causal based visual recognition system, and an explainable AI. 
\vskip-.205in
\section*{Acknowledgement}
\vskip-.05in
The authors are thankful to Ernest Davis for feedback provided during the writing of this document, and to Donald Rubin and Andrew Gelman for helpful discussions.


\newpage
\renewcommand{\baselinestretch}{.95}
{\small
\bibliographystyle{ieee}
\vspace{-.1in}
\bibliography{all,ekrefs}
}

\end{document}